\documentclass{article}
\PassOptionsToPackage{numbers, sort&compress}{natbib}
\usepackage[preprint]{dms}

\usepackage[utf8]{inputenc} 
\usepackage[T1]{fontenc}    
\usepackage{hyperref}       
\usepackage{url}            
\usepackage{booktabs}       
\usepackage{amsfonts}       
\usepackage{nicefrac}       
\usepackage{microtype}      
\usepackage{xcolor}         
\usepackage{subcaption}     
\usepackage{graphicx}       
\usepackage{tabularx}
\usepackage{siunitx}
\usepackage{float}
\usepackage{pifont}

\renewcommand{\b}[1]{\textbf{#1}}

\title{Project Ariadne: Prompt-Conditioned Route Generation for Synthesis Planning}

\author{%
  Anton Morgunov* \\
  Yale University\\
  \texttt{anton@ischemist.com} \\
  \And
  Victor Batista \\
  Yale University\\
  \texttt{victor.batista@yale.edu} \\
}

\begin{document}

\maketitle

\begin{abstract}
Retrosynthetic planning seeks to connect a target molecule to commercially available starting materials through a multistep route.
Classical planners construct such routes by iteratively applying single-step reaction models within a search procedure; constrained variants often require specialized algorithms or architectural changes.
Direct route generation reframes retrosynthesis as sequence generation, but existing direct-generation methods still train separate models for different planning specifications.
We introduce Ariadne, a decoder-only route generator that represents the target, optional constraints, and route in one prompt-completion sequence.
On the RetroCast/PaRoutes \texttt{mkt-cnv-160} benchmark family, one 24-layer checkpoint follows route-depth and required-starting-material prompts: adding the corresponding prompt fields raises Solv-0 by 13.7 points for depth constraints and 31.2 points for required-leaf constraints.
Ariadne also improves over DESP, a bidirectional search planner, on required-leaf Top-10 and Solv-0 in 24 GPU-minutes versus 6.8 GPU-hours.
On standard reconstruction, Ariadne is comparable to DMS Explorer XL at about half the reported inference time.
Across additional target-only benchmarks, Ariadne's clearest gains are on route-holdout reconstruction, whereas AiZynthFinder MCTS remains stronger on several Solv-0 comparisons.
These results extend sequence generation from specialist retrosynthesis models to prompt-conditioned structural route generation.
We release the \href{https://github.com/ischemist/project-ariadne}{codebase and training scripts} to support further work, but do not introduce Tier-1--3 route checkers; those remain the main bottleneck before models of this kind can become useful to experimental chemists.
\end{abstract}

\section{Introduction}

Machine learning is expected to significantly accelerate, if not revolutionize, the often decades-long and billion-dollar process of drug discovery. 
A persistent bottleneck during hit-to-lead and lead optimization stages is a simple question: can this molecule be easily made?~\cite{grandchallenges} 
While significant effort has been put into attempts to answer that question directly by training synthetic accessibility predictors~\cite{sascore_2009,scscore_2018,rascore_2020,syba}, an emerging consensus is that the only truly reliable measure of synthesizability is the explicit construction of a synthesis plan connecting a desired target to a set of commercially available building blocks~\cite{parrot_2024,syntaxofmatter}. 
This synthesis plan can be constructed in either direction: by starting from building blocks through synthesis-aware forward design~\cite{rgfn_2024,synthformer_2025,rxnflow_2025,prexsyn_2025}, or by applying retrosynthetic analysis to the target molecule~\cite{corey_1969}.

The prevalent approach to multistep retrosynthetic planning is built from two components: a \textit{single-step reaction predictor} that is applied iteratively to the target molecule and resulting precursor candidates, and a \textit{search algorithm} prioritizing the most promising branches of the resulting search space~\cite{mcts_2018,dfpn_2019,selfplay_2019,green_mcts_2020,retrostar_2020,aizyn_2020,grasp_2022,retrograph_2022,egmcts_2023,liu_2023,evoretro_2023,meea_2024,retrofallback_2024,resynz_2024,enhmcts_2025,dreamretroer_2025,higherlev_2025,synplanner_2025,treemdp_2025}.
Hybrid systems keep explicit search but add learned, retrieval-based, or language-model guidance to steer expansion and pruning~\cite{retrek_2022,nested_mcts_2024,llamole_2024,desp_2024,chimera_2025,larc_2025,synthelite_2025,song_2025,retrogfn_2025}.
An emerging alternative is direct generation of the synthesis plan represented as a single string~\cite{moltransformer_2020,autosynroute_2020,multistepttl_2023,dms_2025,andronov_2025,synllama_2025,tempre_2025,llmmulti_2025,retrosynformer_2026}. 
For example, \citeauthor{dms_2025} trained a series of encoder-decoder transformers to "translate" a SMILES specification of the target compound into a stringified (via depth-first search) representation of the multistep route. 
These DirectMultiStep models were also extended to constrained versions of retrosynthetic planning, such as finding a route with a specified starting-material structure or desired route depth, but each such problem required training a specialist model.

In this work, we extend the DirectMultiStep sequence formulation from separately trained encoder-decoder models to a single decoder-only task language for route generation.
Ariadne represents the target, optional planning constraints, and route in one sequence, so the same checkpoint can be queried with different task specifications at inference time.
As a proof of concept, we study target-only reconstruction together with route-depth and required starting-material prompts.
We evaluate these outputs within the existing Solv-N and RetroCast framework, using route reconstruction and constraint-aware Solv-0 to test whether generated routes satisfy benchmark specifications~\cite{syntaxofmatter,paroutes,retrocast_2025}.
These metrics evaluate the structural route plan: the reaction topology, stock termination, and prompt-specified constraints.
Direct experimental use would require additional quantitative planning layers, such as reaction plausibility assessment, condition prediction, procedure generation, and higher-tier executability checks discussed in the Solv-N framework~\cite{syntaxofmatter}.

\section{Preliminaries}
\subsection{Definitions}
A retrosynthetic \textit{route} is a sequence of \textit{reactions} working backward from a \textit{target molecule} to a set of proposed \textit{starting materials} (or \textit{leaves}). All models discussed herein inherit ambiguities from their patent-derived training data, which may not distinguish core \textit{reactants} from auxiliary \textit{reagents} and may omit reaction conditions (e.g., solvent, temperature). A predicted route is therefore not a complete experimental protocol but a high-level topological plan, the validity of which rests on an unevaluated assumption that viable conditions exist for each transformation.

\subsection{Evaluation}
We distinguish the \textit{generation prompt}, which is supplied to Ariadne before decoding, from the \textit{scoring task}, which defines the constraints used by RetroCast during evaluation.
This distinction lets us ask, for example, how target-only generations score under the stricter required-leaf task, or how required-leaf prompts behave when scored under the standard target-only task.

We report two complementary metric sets. First, we report Tier-0 validity, which is the share of targets that have at least one route where all reactions are Tier-0 valid, and Solv-0, which is the share of targets that have at least one Tier-0-valid route that satisfies the scoring task constraints~\cite{syntaxofmatter}. For \texttt{mkt-cnv-160}, the scoring task constraint is simply termination in the ASKCOS Buyables stock of commercially available compounds~\cite{askcos_2025,higherlev_2025}. For \texttt{mkt-cnv-160-leaf}, the scoring task constraint is stock termination together with the presence of a specified starting material among the leaves. For \texttt{mkt-cnv-160-depth}, it is stock termination together with the requested route depth.
Additional target-only benchmarks use the same RetroCast convention: \texttt{mkt-} benchmarks are scored with ASKCOS Buyables, whereas \texttt{ref-} benchmarks are scored with the patent-derived PaRoutes stocks distributed with the benchmark definitions.

In the absence of established Tier-1--3 validity checking protocols, and following the proposed separation of method development from introduction of new evaluation metrics~\cite{retrocast_2025,syntaxofmatter}, we report benchmark route reconstruction as a proxy metric of route quality. We use the standard RetroCast implementation of scoring and report Top-$K$ accuracy, that is, whether a reference route was produced within the first $K$ candidates.

\subsection{Data Representation}
\begin{figure}[t]
\centering
\includegraphics[width=\textwidth]{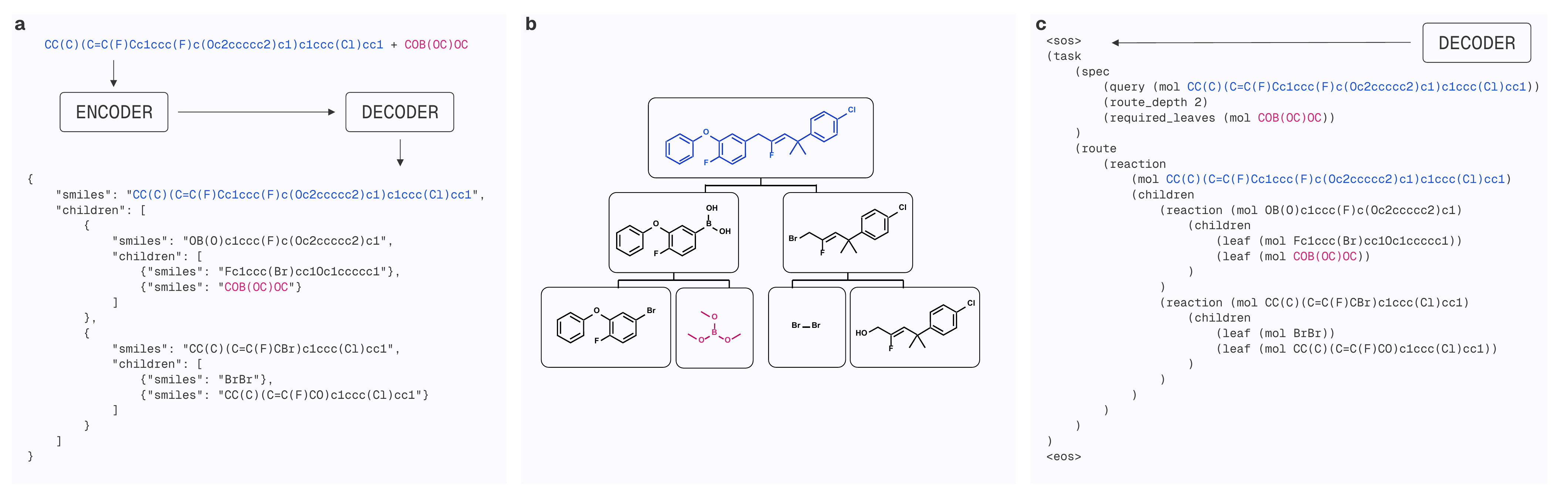}
\caption{Data representation shift from DirectMultiStep to Ariadne. \textbf{(a)} DirectMultiStep treats multistep retrosynthesis as sequence translation from target molecule (and optionally appended constraints) to a synthesis plan. \textbf{(b)} The skeletal structure of the route encoded in (a) and (c). \textbf{(c)} Ariadne represents the same problem as one decoder-only task sequence containing both the prompt-side specification and the route-side answer.}
\label{fig:data-representation}
\end{figure}

Ariadne is a decoder-only transformer trained on stringified representations of synthesis planning tasks. Each training example is a rooted S-expression (see Fig.~\ref{fig:data-representation}) with two parts: a problem specification and the route that solves it:
\begin{quote}
\texttt{(task (spec ...) (route ...))}
\end{quote}
where \texttt{spec} contains the prompt-side information and \texttt{route} contains the target route tree. The \texttt{spec} contains a target molecule represented as \texttt{(query (mol ...))} and any optional constraints. A route is represented recursively. A leaf node is written as \texttt{(leaf (mol ...))}. A reaction node is written as \texttt{(reaction (mol ...)\allowbreak{} (children ...))}, where the children are precursor routes.

The same route can be converted into different training sequences by changing only the \texttt{spec} block. In the simplest target-only mode, denoted \texttt{T}, the specification contains only the target molecule. 
In \texttt{TL}, it also contains \texttt{route\_depth}. 
In \texttt{TSd}, it contains one required starting material, chosen from the deepest leaf of the route. 
In \texttt{TLSd}, it contains both route depth and that required starting material. 
During training we also generate \texttt{TSe} and \texttt{TLSe} sequences in which the required-leaf field is instantiated once for each route leaf. Here \texttt{d} denotes the deepest leaf and \texttt{e} denotes enumeration over leaves; the \texttt{TSd} and \texttt{TLSd} evaluation prompts select the deepest-leaf instance from the corresponding enumerated training variants. 
The field is named \texttt{required\_leaves} because the representation supports multiple required starting materials, but all constrained experiments in this work use one required starting material. 
These variants let one model see the same route distribution under different amounts of information and naturally offset underrepresentation of longer routes: a longer route typically has more leaves and therefore contributes more leaf-conditioned sequences.

Training data also augments task sequences by permuting sibling order in the route section. 
We generate three deterministic permutations: one that permutes children recursively, one that permutes only the root children, and one that permutes the deepest branching node.

We tokenize this representation with a small S-expression-aware tokenizer. 
Parentheses and structural labels such as \texttt{task}, \texttt{spec}, \texttt{query}, \texttt{route}, \texttt{reaction}, \texttt{leaf}, \texttt{children}, \texttt{route\_depth}, and \texttt{required\_leaves} are atomic tokens. 
SMILES are then split character-wise. 
During training, the prompt portion is masked out of the loss; the model is trained to generate the route side of the sequence conditioned on the specification.

\subsection{Training Data}

All Ariadne models were trained on the \texttt{v2026-05-12} canonical split of the PaRoutes dataset preprocessed with RetroCast. RetroCast provides two versions: a \textit{route} holdout, which guarantees that no route from the test set is represented as-is in the training set, and a \textit{reaction} holdout, which guarantees that no test-route reaction appears in any training route after RetroCast canonicalization. 
\texttt{v2026-05-12-route} is equivalent to the training set used in the original DirectMultiStep work. \texttt{v2026-05-12-reaction} follows the stricter filtering proposed by \citeauthor{tempre_2025}, who argued that route-based filtering leads to unfair data leakage.

\section{Results and Discussion}

DirectMultiStep showed that target-only and constrained retrosynthesis can be written as sequence-to-sequence problems, but each task variant (e.g. unidirectional and bidirectional search) required a separate model.
Ariadne pushes that idea one step further: the target and optional constraints are prompt fields, and a single route generator completes the prompt with a synthesis tree.
The \texttt{mkt-cnv-160} benchmark family gives a controlled test of this interface because it keeps the same 160 targets while changing only the route constraints.
The base task requires stock termination in ASKCOS Buyables; the depth variant additionally fixes route depth, and the leaf variant additionally fixes one starting material that must appear among the route leaves.

\subsection{One checkpoint handles multiple planning specifications}

Table~\ref{tab:mkt-cnv-160-constrained-prompts} isolates constraint following by holding the Ariadne checkpoint and generation procedure fixed while changing the generation prompt.
The target-only \texttt{T} rows generate unconstrained candidates and score them under the stricter depth or leaf scoring tasks, giving the baseline rate at which generation already satisfies the added requirement.
The constrained rows include the corresponding field in the generation prompt before decoding.
On the depth benchmark, adding the requested depth with the \texttt{TL} prompt raises Solv-0 from 76.9\% to 90.6\%.
On the required-leaf benchmark, adding the benchmark-specified required starting material with the \texttt{TSd} prompt raises Solv-0 from 50.0\% to 81.2\% and Top-10 reconstruction from 26.2\% to 37.5\%.

The required-leaf benchmark also gives a direct comparison to DESP~\cite{desp_2024}, a bidirectional search planner built for target-plus-starting-material constraints.
With the \texttt{TSd} prompt, Ariadne reaches 37.5\% Top-10 and 81.2\% Solv-0 in 24 GPU-minutes.
The best DESP setting reaches 17.5\% Top-10 and 71.2\% Solv-0 in 6.8 GPU-hours.
A paired bootstrap comparison on Top-10 gives Ariadne a +20.0-point advantage, with 95\% CI [10.6, 28.8].
Ariadne therefore improves both Top-10 and Solv-0 while using about $17\times$ less GPU time.

\begin{table}[t]
\centering
\scriptsize
\resizebox{\textwidth}{!}{%
\begin{tabular}{l l l r r r r r r}
\toprule
Scoring task & Model & Generation prompt / setting & Tier-0 valid & Solv-0 & Top-1 & Top-10 & Top-10 CI & Time \\
\midrule
\texttt{depth} & Ariadne 24L & \texttt{T} & 100.0\% & 76.9\% & 23.8\% & 42.5\% & [35.0, 50.0] & 19.4 min \\
\texttt{depth} & Ariadne 24L & \texttt{TL} & 100.0\% & 90.6\% & 26.9\% & \b{46.2\%} & [38.8, 53.8] & 19.3 min \\
\addlinespace
\texttt{leaf} & DESP \texttt{retro\_sd} & 100 iter & 87.5\% & 49.4\% & 12.5\% & 13.8\% & [8.8, 19.4] & 1.1 hr \\
\texttt{leaf} & DESP \texttt{retro\_sd} & 500 iter & 97.5\% & 60.0\% & 15.0\% & 16.2\% & [10.6, 21.9] & 5.5 hr \\
\texttt{leaf} & DESP \texttt{f2e} & 100 iter & 91.9\% & 61.9\% & 15.6\% & 16.9\% & [11.2, 23.1] & 1.4 hr \\
\texttt{leaf} & DESP \texttt{f2e} & 500 iter & 98.8\% & 71.2\% & 16.2\% & 17.5\% & [11.9, 23.8] & 6.8 hr \\
\texttt{leaf} & Ariadne 24L & \texttt{T} & 100.0\% & 50.0\% & 13.8\% & 26.2\% & [19.4, 33.1] & 19.4 min \\
\texttt{leaf} & Ariadne 24L & \texttt{TSd} & 96.2\% & 81.2\% & 18.8\% & \b{37.5\%} & [30.0, 45.0] & 24.0 min \\
\bottomrule
\end{tabular}}
\caption{Constraint-aware evaluation on the \texttt{mkt-cnv-160-depth} and \texttt{mkt-cnv-160-leaf} scoring tasks. Ariadne rows use the 24-layer model trained on the \texttt{v2026-05-12-reaction} split at the 14B-token checkpoint with beam size 50. \texttt{T} is target-only prompting, \texttt{TL} adds the requested route depth, and \texttt{TSd} adds the benchmark-specified required starting material to the generation prompt. Top-10 CI gives the 95\% bootstrap interval.}
\label{tab:mkt-cnv-160-constrained-prompts}
\end{table}

\subsection{The unified model preserves standard route reconstruction}

We also evaluate whether the unified planner remains competitive on the standard target-only task.
Table~\ref{tab:mkt-cnv-160-main-results} reports public SynthArena baselines together with local runs on the matched \texttt{v2026-05-12} training splits.
On the stricter reaction holdout, Ariadne 24L is comparable to retrained MCTS on route reconstruction: 22.5\% versus 19.4\% Top-1 and 42.5\% versus 35.0\% Top-10 (paired bootstrap 95\% CI for the Top-10 difference: [-0.6, 15.6]).
MCTS remains stronger on Solv-0 (92.5\% versus 81.2\%) and faster at inference (10.9 versus 19.4 minutes).
On the route holdout, which is the closest comparison to the original DirectMultiStep split, Ariadne 24L (15.8 M parameters) reaches 38.1\% Top-1 and 59.4\% Top-10, which is comparable to the DMS Explorer XL (50 M parameters) Top-10 result of 57.5\% while reducing generation time from 47.6 to 23.9 minutes.

\begin{table}[t]
\centering
\small
\begin{tabularx}{\textwidth}{p{0.28\textwidth}r r r r r r}
\toprule
Model & Tier-0 valid & Solv-0 & Top-1 & Top-10 & Top-10 CI & Time \\
\midrule
\multicolumn{7}{l}{\textit{Reprinted from RetroCast/SynthArena}} \\
AiZynthFinder MCTS & -- & 98.1\% & 21.3\% & 41.3\% & -- & 17.8 min \\
DMS Explorer XL & -- & 96.3\% & 33.8\% & \b{57.5\%} & -- & 47.6 min \\
\addlinespace
\multicolumn{7}{l}{\textit{Training on \texttt{v2026-05-12-reaction}}} \\
AiZynthFinder MCTS & 100.0\% & 92.5\% & 19.4\% & 35.0\% & [28.1, 42.5] & 10.9 min \\
Ariadne 12L & 100.0\% & 74.4\% & 15.6\% & 31.9\% & [25.0, 39.4] & 10.1 min \\
Ariadne 24L & 100.0\% & 81.2\% & 22.5\% & \b{42.5\%} & [35.0, 50.0] & 19.4 min \\
\addlinespace
\multicolumn{7}{l}{\textit{Training on \texttt{v2026-05-12-route}}} \\
AiZynthFinder MCTS & 100.0\% & 93.1\% & 23.8\% & 40.6\% & [33.1, 48.1] & 10.5 min \\
Ariadne 12L & 100.0\% & 87.5\% & 27.5\% & 47.5\% & [40.0, 55.0] & 11.8 min \\
Ariadne 24L & 100.0\% & 95.6\% & 38.1\% & \b{59.4\%} & [51.9, 66.9] & 23.9 min \\
\bottomrule
\end{tabularx}
\caption{\texttt{mkt-cnv-160} Top-$K$ reconstruction and Solv-0 (stock termination) results. Public AiZynthFinder MCTS and DMS Explorer XL values are taken from the \href{https://syntharena.ischemist.com/leaderboard?benchmarkId=cmisc0flu0000boddjstwifeo}{SynthArena leaderboard}. Ariadne 12L rows use the 20B-token checkpoint, and Ariadne 24L rows use the 14B-token checkpoint. AiZynthFinder runs use 100 MCTS iterations and maximum search depth 6. Top-10 CI gives the 95\% bootstrap interval for newly evaluated rows.}
\label{tab:mkt-cnv-160-main-results}
\end{table}

\subsection{Route and reaction holdouts separate chemistry generalization from planning ability}

Table~\ref{tab:mkt-cnv-160-main-results} shows how the same planners behave on the reaction and route holdouts, giving two complementary views of reaction coverage and route assembly.
The reaction holdout removes every reaction from the benchmark routes, so success is a stricter proxy for both route planning and generalization beyond the exact single-step reactions present in training.
The route holdout removes exact benchmark routes but can leave their component reactions (if they're present in other routes) in training, giving a cleaner read on whether a planner can assemble covered transformations into the reference route.

Moving from the reaction holdout to the route holdout, MCTS Top-10 rises modestly from 35.0\% to 40.6\%, whereas Ariadne 24L rises from 42.5\% to 59.4\%.
On the route holdout itself, the paired Top-10 comparison gives Ariadne a +18.8-point advantage over MCTS, with 95\% CI [10.6, 26.9].
On \texttt{mkt-cnv-160}, these comparisons indicate that additional single-step reaction coverage translates into larger route-level gains for Ariadne than for MCTS.
The same trend appears in Solv-0, which rises from 92.5\% to 93.1\% for MCTS and from 81.2\% to 95.6\% for Ariadne 24L.
The additional commercial-stock benchmark gives the same picture: on \texttt{mkt-lin-500}, Ariadne improves Top-10 on both holdouts and all route-holdout reconstruction metrics, while MCTS remains a strong Solv-0 baseline (Table~\ref{tab:si-mkt-lin-500-results}).

\subsection{Reference-stock benchmarks expose stock-conditioning limits}

A current limitation of Ariadne is that generation cannot yet be conditioned on an arbitrary user-specified stock set.
Such stock control is central to practical planning, where the useful terminal set may include in-house building blocks in addition to commercially available compounds.
Search-based planners can enforce this constraint during expansion; the present Ariadne model can only be filtered against the desired stock after generation.
The \texttt{ref-*} benchmarks expose this limitation by replacing ASKCOS Buyables with PaRoutes reference stocks: \texttt{ref-cnv-400} and \texttt{ref-lin-600} use \texttt{n5}, and \texttt{ref-lng-84} uses combined \texttt{n1}/\texttt{n5} stocks (Tables~\ref{tab:si-ref-cnv-400-results}--\ref{tab:si-ref-lng-84-results}).

On route holdouts, Ariadne reaches 68.2\% Solv-0 on \texttt{ref-cnv-400} and 74.8\% on \texttt{ref-lin-600}, below the MCTS values of 85.2\% and 83.2\%.
The reaction holdouts widen the gap: 33.0\% versus 76.5\% on \texttt{ref-cnv-400}, and 55.8\% versus 78.8\% on \texttt{ref-lin-600}.
Although Ariadne remains competitive on route-holdout Top-10 reconstruction, these results indicate that arbitrary-stock conditioning is an important area for future research.

\subsection{Larger models recover deeper routes more often}

Table~\ref{tab:mkt-cnv-160-depth-results} shows that scaling Ariadne improves reconstruction of deeper routes, especially on the route holdout.
On the reaction holdout, deeper routes remain difficult for every method: MCTS falls from 57.5\% at depth 2 to 17.5\% at depth 5, and Ariadne 24L falls from 67.5\% to 20.0\%.
On the route holdout, the profile changes more substantially for Ariadne than for MCTS.
MCTS still declines with depth, while Ariadne 24L reaches 45.0\% Top-10 at depth 4 and 52.5\% at depth 5.
Within the limits of the 40-target strata, this is consistent with additional scale helping Ariadne turn better reaction coverage into longer-range route reconstruction.

\begin{table}[t]
\centering
\scriptsize
\resizebox{\textwidth}{!}{%
\begin{tabular}{l r r r r}
\toprule
Model & Depth 2 & Depth 3 & Depth 4 & Depth 5 \\
\midrule
\multicolumn{5}{l}{\textit{Training on \texttt{v2026-05-12-reaction}}} \\
AiZynthFinder MCTS & 57.5\% [42.5, 72.5] & 42.5\% [27.5, 57.5] & 22.5\% [10.0, 35.0] & 17.5\% [7.5, 30.0] \\
Ariadne 12L & 47.5\% [32.5, 62.5] & 40.0\% [25.0, 55.0] & 20.0\% [7.5, 32.5] & \b{20.0\% [7.5, 32.5]} \\
Ariadne 24L & \b{67.5\% [52.5, 82.5]} & \b{52.5\% [37.5, 67.5]} & \b{30.0\% [17.5, 45.0]} & \b{20.0\% [7.5, 32.5]} \\
\addlinespace
\multicolumn{5}{l}{\textit{Training on \texttt{v2026-05-12-route}}} \\
AiZynthFinder MCTS & 60.0\% [45.0, 75.0] & 45.0\% [30.0, 60.0] & 30.0\% [17.5, 45.0] & 27.5\% [15.0, 42.5] \\
Ariadne 12L & 57.5\% [42.5, 72.5] & 50.0\% [35.0, 65.0] & 32.5\% [17.5, 47.5] & 50.0\% [35.0, 65.0] \\
Ariadne 24L & \b{75.0\% [60.0, 87.5]} & \b{65.0\% [50.0, 80.0]} & \b{45.0\% [30.0, 60.0]} & \b{52.5\% [37.5, 67.5]} \\
\bottomrule
\end{tabular}}
\caption{\texttt{mkt-cnv-160} Top-10 route reconstruction accuracy by reference route depth for local runs. Ariadne 12L rows use the 20B-token checkpoint, and Ariadne 24L rows use the 14B-token checkpoint. Each depth stratum contains 40 targets. Values are reported as mean with 95\% bootstrap interval.}
\label{tab:mkt-cnv-160-depth-results}
\end{table}

\subsection{The first disconnection is the main bottleneck}

Table~\ref{tab:mkt-cnv-160-reconstruction-diagnostics} separates recovery of the root reaction from recovery of the full route conditional on getting the root right.
For MCTS, these quantities are similar; on the route split, both root Top-10 and \texttt{route$\mid$root} Top-10 are 63.7\%.
In contrast, for Ariadne \texttt{route$\mid$root} Top-10 is consistently higher than root Top-10, with the largest gaps on the reaction split and for smaller models.
For example, Ariadne 12L on the reaction split reaches 46.9\% root Top-10 but 68.0\% \texttt{route$\mid$root} Top-10, indicating that the main limitation is choosing the first disconnection rather than completing the route once that choice is correct.

\begin{table}[t]
\centering
\scriptsize
\resizebox{\textwidth}{!}{%
\begin{tabular}{l r r r r r r r r}
\toprule
Model & Root T1 & Route$\mid$root T1 & Root T10 & Route$\mid$root T10 & Prefix-1 T10 & Prefix-2 T10 & Prefix-3 T10 & Distinct roots \\
\midrule
\multicolumn{9}{l}{\textit{Training on \texttt{v2026-05-12-reaction}}} \\
AiZynthFinder MCTS & 41.2\% & 47.0\% & 62.5\% & 56.0\% & 62.5\% & 40.0\% & 35.0\% & 2.831 \\
Ariadne 12L & 28.1\% & 55.6\% & 46.9\% & 68.0\% & 46.9\% & 34.4\% & 31.2\% & 2.056 \\
Ariadne 24L & 48.1\% & 46.8\% & 60.0\% & 70.8\% & 60.0\% & 46.9\% & 43.1\% & 2.075 \\
\addlinespace
\multicolumn{9}{l}{\textit{Training on \texttt{v2026-05-12-route}}} \\
AiZynthFinder MCTS & 44.4\% & 53.5\% & 63.7\% & 63.7\% & 63.7\% & 46.9\% & 41.9\% & 2.950 \\
Ariadne 12L & 38.8\% & 71.0\% & 57.5\% & 82.6\% & 57.5\% & 50.0\% & 48.8\% & 2.444 \\
Ariadne 24L & 54.4\% & 70.1\% & 70.0\% & 84.8\% & 70.0\% & 61.3\% & 60.6\% & 2.631 \\
\bottomrule
\end{tabular}}
\caption{\texttt{mkt-cnv-160} reconstruction diagnostics. Ariadne 12L rows use the 20B-token checkpoint, and Ariadne 24L rows use the 14B-token checkpoint. Root columns measure recovery of the reference root reaction at the indicated $K$. Route$\mid$root columns measure full-route recovery among targets whose root reaction was recovered at the same $K$. Prefix columns report Top-10 reconstruction of reference route prefixes of depth 1, 2, and 3. Mean distinct roots is computed over the Top-10 candidates.}
\label{tab:mkt-cnv-160-reconstruction-diagnostics}
\end{table}

\subsection{Beam search mainly improves root selection}

To interpret the beam-50 results used above, we sweep beam width on \texttt{mkt-lin-500} and ask whether larger beams improve route completion itself or mainly increase the chance of sampling the correct first disconnection (Table~\ref{tab:si-mkt-lin-500-beam-sweep}).
Increasing the beam from 1 to 50 raises root Top-10 from 17.4\% to 63.0\% on the reaction holdout and from 22.0\% to 72.4\% on the route holdout.
Conditional route recovery changes much less: \texttt{route$\mid$root} Top-10 remains between 57.5\% and 64.4\% on the reaction holdout and between 70.1\% and 72.7\% on the route holdout.
The gains in Solv-0 and Top-10 reconstruction therefore come mainly from improved root selection; on the reaction holdout, Solv-0 increases from 33.8\% to 90.4\% and Top-10 increases from 10.0\% to 37.2\%.
Prompt fields that improve root selection, or dynamic beam allocation based on target complexity or search progress, may reduce decoding time while preserving high Solv-0.
Developing such decoding strategies is an area for future work.

\subsection{Prefix-LM attention does not improve prompted generation}

DirectMultiStep used an encoder-decoder architecture, so the route decoder attended to a bidirectional representation of the complete task specification.
This makes it natural to ask whether Ariadne should recover that property with Prefix-LM attention~\cite{prefixlm_2019}, which allows bidirectional attention within the prompt while keeping route generation causal.
Table~\ref{tab:mkt-cnv-160-prefix-causal} shows no overall benefit from restoring bidirectional attention within the prompt.
In the 24-layer \texttt{T-TL-TLSe-TSe} comparison, fully causal attention gives higher Top-1, Top-10, and root-recovery values for every evaluated prompt.
It also gives higher \texttt{route$\mid$root} Top-10 for all three evaluated prompts.
We therefore use the simpler fully causal attention mask for Ariadne.

\begin{table}[t]
\centering
\scriptsize
\resizebox{\textwidth}{!}{%
\begin{tabular}{l l r r r r r r r}
\toprule
Attention & Prompt & Tier-0 valid & Solv-0 & Top-1 & Top-10 & Top-10 CI & Root T10 & Route$\mid$root T10 \\
\midrule
\multicolumn{9}{l}{\textit{Training prompts: \texttt{T-TL-TLSe-TSe}; 24-layer model}} \\
causal & \texttt{T} & 100.0\% & 81.2\% & 22.5\% & \b{42.5\%} & [35.0, 50.0] & 60.0\% & 70.8\% \\
causal & \texttt{TL} & 100.0\% & 90.6\% & 26.9\% & \b{46.2\%} & [38.8, 53.8] & 66.9\% & 69.2\% \\
causal & \texttt{TSd} & 96.2\% & 81.2\% & 18.8\% & \b{37.5\%} & [30.0, 45.0] & 55.6\% & 67.4\% \\
\addlinespace
prefix & \texttt{T} & 100.0\% & 80.6\% & 13.8\% & 29.4\% & [22.5, 36.3] & 43.8\% & 67.1\% \\
prefix & \texttt{TL} & 99.4\% & 86.9\% & 18.8\% & 33.8\% & [26.9, 41.2] & 50.6\% & 66.7\% \\
prefix & \texttt{TSd} & 85.0\% & 70.0\% & 7.5\% & 18.1\% & [12.5, 24.4] & 28.1\% & 64.4\% \\
\bottomrule
\end{tabular}}
\caption{\texttt{mkt-cnv-160} attention-mask ablation using standard stock-termination scoring. Causal models use ordinary left-to-right attention for all tokens. Prefix models allow bidirectional attention within the generation prompt while keeping route generation causal. Here \texttt{TL} and \texttt{TSd} add prompt fields before decoding, but RetroCast scoring still uses the standard stock-termination task rather than the constrained depth or leaf scoring tasks in Table~\ref{tab:mkt-cnv-160-constrained-prompts}. Top-10 CI gives the 95\% bootstrap interval. For the causal \texttt{TL} and \texttt{TSd} rows, Solv-0 coincides with the corresponding constrained rows in Table~\ref{tab:mkt-cnv-160-constrained-prompts} because the stock-terminated successes also satisfy the prompted constraint.}
\label{tab:mkt-cnv-160-prefix-causal}
\end{table}

\section{Conclusion}
Ariadne shows that direct multistep route generation can be formulated as prompt-conditioned decoding rather than as a collection of separately trained specialist models.
In one 24-layer checkpoint, the same target/constraint/route language supports target-only reconstruction, route-depth prompts, and required-starting-material prompts.
On the \texttt{mkt-cnv-160} benchmark family, this checkpoint follows the added constraint fields, remains competitive with DirectMultiStep Explorer XL on standard reconstruction, and exceeds the evaluated DESP settings on required-leaf Top-10 and Solv-0 while using 17x less GPU time.

Whenever route termination depends on in-house compounds rather than off-the-shelf buyables, a planner must be able to condition on an arbitrary stock set.
The \texttt{ref-*} benchmarks instantiate this setting by replacing ASKCOS Buyables with custom stocks derived from reference-route leaves.
MCTS receives those stocks during search and correspondingly reaches 85.2\% Solv-0 on the \texttt{ref-cnv-400} route holdout.
Because the present Ariadne model can only be filtered against those stocks after decoding, its Solv-0 on the same task is only 68.2\%.

To understand why Ariadne is competitive on reference-route reconstruction, we report diagnostics that separate recovery of the first disconnection from recovery of the rest of the route.
These diagnostics show that Ariadne often reconstructs the rest of the route correctly once the first disconnection is recovered.
The beam-width sweep on \texttt{mkt-lin-500} points to the same mechanism: increasing beam width mainly increases the probability of recovering that first disconnection, which then raises root-reaction Top-10, Solv-0, and Top-10 reconstruction.
Together with the stock-conditioning results above, this identifies stock-aware prompting and more reliable first-disconnection selection as important areas for further research.

\section{Outlook}
This work should not be interpreted as an attempt to replace MCTS or explicit search more generally.
Rather, in the spirit of the bitter lesson~\cite{bitter_lesson,syntaxofmatter}, explicit search and direct generation appear to offer complementary scaling axes: search provides direct control over stock termination and other hard constraints, while learned generators can amortize recurring route structure into faster prompt-conditioned inference.
One natural next step is therefore to use search to generate constraint-satisfying trajectories and distill those trajectories into direct generators.
For Ariadne specifically, the immediate model-side extensions are arbitrary-stock prompting, prompt fields or decoding policies that improve root selection, dynamic beam allocation based on target complexity or search progress, and speculative decoding~\cite{andronov_2025}.
On the evaluation side, Solv-0 and Top-$K$ reconstruction evaluate structural route plans under Tier-0 checks, so the field needs standardized Tier-1--3 validation protocols before performance on these benchmarks can be translated into claims about experimentally executable synthesis.
We release the codebase and full training scripts so that Ariadne can serve as a reproducible baseline for these directions.

\section{Implementation Details}

\subsection{Model}
In addition to switching from an encoder-decoder setup in DirectMultiStep to a decoder-only architecture for Ariadne, we also updated the implementation of transformer blocks (see Table~\ref{tab:dms-ariadne-comparison}).

The implemented model is a pre-normalization decoder with token embeddings, RMSNorm, rotary position embeddings, multi-head self-attention, and SwiGLU feed-forward blocks. 
The runs reported here use dense feed-forward layers. 
The code also supports sparse mixture-of-experts blocks, but those are not used for the main results. 
We vary model scale mainly by layer count while holding hidden size at 256 and using eight attention heads.
This results in 7.9 M parameters for the 12 layer model and 15.8 M for the 24 layer model.
For reference, DMS Explorer XL has 50 M parameters~\cite{dms_2025}.

\begin{table}[ht]
\centering
\small
\begin{tabularx}{\textwidth}{p{0.22\textwidth}X X}
\toprule
Choice & DirectMultiStep & Ariadne \\
\midrule
Architecture & Encoder-decoder translation~\cite{transformer,dms_2025}. & Decoder-only language modeling. \\
\addlinespace
Position embedding & Learned absolute embeddings~\cite{transformer}. & RoPE~\cite{rope}. \\
\addlinespace
Normalization & LayerNorm~\cite{layernorm}. & RMSNorm~\cite{rmsnorm}. \\
\addlinespace
Normalization order & Post-LN residual blocks~\cite{transformer}. & Pre-LN residual blocks~\cite{preln}. \\
\addlinespace
Feed-forward block & ReLU/GELU MLP~\cite{transformer,gelu}. & SwiGLU MLP~\cite{swiglu,llama}. \\
\addlinespace
Optimizer & AdamW~\cite{adamw}. & Muon with Moonshot scaling for dense matrices, AdamW elsewhere~\cite{muon,moonshot_muon,adamw}. \\
\bottomrule
\end{tabularx}
\caption{Main implementation differences between DirectMultiStep and Ariadne.}
\label{tab:dms-ariadne-comparison}
\end{table}

\subsection{Training}
The supervised objective is next-token cross entropy on the route part of the sequence. 
Prompt tokens and padding tokens are masked out of the labels. 
Training uses Hugging Face Accelerate for device placement, gradient accumulation, checkpointing, and mixed precision. 
The main runs use one GPU per run with bfloat16 autocast on supported accelerators. 
Dense two-dimensional transformer weights are optimized with Muon using Moonshot update scaling. 
Embeddings, output head, normalization parameters, and other residual parameters use AdamW-style groups. 
Gradients are clipped to unit norm. Learning-rate schedules, logging, validation, checkpointing, and budget accounting are token-based rather than epoch-based. 
We use length bucketing to reduce computational costs associated with excessive padding, resulting in a roughly $12\times$ speedup.

\subsection{Generation}
Generation uses deterministic batched beam search with a KV cache~\cite{efficient_inference}. 
The prompt is the task specification plus the start of the route wrapper and ends immediately after the \texttt{children} token in that incomplete wrapper. 
The prompt is run once, cached key/value states are expanded across beams, and subsequent decoding steps feed only the newest token. 
After each beam update, the cache is reordered to match the surviving beams. 
All reported Ariadne generations use beam size 50, length penalty 0.5, and a 1200-token generation limit; decoding stops earlier when all beams emit \texttt{<eos>}.

\subsection{Evaluation}
We use the standard RetroCast v0.7.x implementations of Solv-N and route reconstruction scoring defined in Preliminaries~\cite{retrocast_2025}. 
Ariadne supplies raw generated candidate routes, but parsing, canonicalization, constraint filtering, duplicate removal, and Top-$K$ reconstruction are all performed by the RetroCast scoring pipeline~\cite{retrocast_2025}. 
Failed parses are preserved as failed candidate slots, so they fail Tier-0 validity and cannot contribute to Solv-0 or reconstruction. 

DMS Explorer XL results in Table~\ref{tab:mkt-cnv-160-main-results} are reprinted from RetroCast/SynthArena, where sequence-based DirectMultiStep runs were performed on Lambda Labs NVIDIA A100 40 GB GPUs~\cite{retrocast_2025}. 
For comparability, Ariadne and DESP evaluations in this work were run on separate clean single-GPU Lambda Labs A100 40 GB instances. 
The MCTS rows use AWS EC2 \texttt{c7i.xlarge} CPU instances, matching the RetroCast/SynthArena protocol for search-based planners~\cite{retrocast_2025}. 
Reported times are planning or generation wall-clock times for the corresponding planner runs.

\section{Data and Software Availability}
Code for processing the dataset, implementing the model architecture, and running training, generation, and evaluation is available under the MIT License at \url{https://github.com/ischemist/project-ariadne}.

\section*{Conflict of Interest}
The authors declare no conflict of interest.

\section{Acknowledgments}

The authors acknowledge a generous allocation of high-performance computing time from NERSC. The development of the methodology was supported by the NSF CCI grant (VSB, Award Number 2124511).  This research was also supported in part by Lambda, Inc.

\medskip
{
\small
\bibliography{reference}

@article{dms_2025,
author = {Shee, Yu and Morgunov, Anton and Li, Haote and Batista, Victor S.},
title = {{DirectMultiStep}: Direct Route Generation for Multistep Retrosynthesis},
journal = {Journal of Chemical Information and Modeling},
volume = {65},
number = {8},
pages = {3903-3914},
year = {2025},
doi = {10.1021/acs.jcim.4c01982},
}

@misc{retrocast_2025,
      title={Procrustean Bed for {AI}-Driven Retrosynthesis: A Unified Framework for Reproducible Evaluation}, 
      author={Anton Morgunov and Victor S. Batista},
      year={2025},
      eprint={2512.07079},
      archivePrefix={arXiv},
      primaryClass={cs.LG},
      url={https://arxiv.org/abs/2512.07079}, 
}

@inproceedings{transformer,
 author = {Vaswani, Ashish and Shazeer, Noam and Parmar, Niki and Uszkoreit, Jakob and Jones, Llion and Gomez, Aidan N and Kaiser, {\L}ukasz and Polosukhin, Illia},
 booktitle = {Advances in Neural Information Processing Systems},
 editor = {I. Guyon and U. Von Luxburg and S. Bengio and H. Wallach and R. Fergus and S. Vishwanathan and R. Garnett},
 pages = {},
 publisher = {Curran Associates, Inc.},
 title = {Attention is All you Need},
 url = {https://proceedings.neurips.cc/paper_files/paper/2017/file/3f5ee243547dee91fbd053c1c4a845aa-Paper.pdf},
 volume = {30},
 year = {2017}
}

@misc{layernorm,
      title={Layer Normalization},
      author={Jimmy Lei Ba and Jamie Ryan Kiros and Geoffrey E. Hinton},
      year={2016},
      eprint={1607.06450},
      archivePrefix={arXiv},
      primaryClass={stat.ML},
      url={https://arxiv.org/abs/1607.06450},
}

@inproceedings{rmsnorm,
 author = {Zhang, Biao and Sennrich, Rico},
 booktitle = {Advances in Neural Information Processing Systems},
 editor = {H. Wallach and H. Larochelle and A. Beygelzimer and F. d\textquotesingle Alch\'{e}-Buc and E. Fox and R. Garnett},
 pages = {},
 publisher = {Curran Associates, Inc.},
 title = {Root Mean Square Layer Normalization},
 url = {https://proceedings.neurips.cc/paper_files/paper/2019/file/1e8a19426224ca89e83cef47f1e7f53b-Paper.pdf},
 volume = {32},
 year = {2019}
}

@inproceedings{preln,
  author       = {Ruibin Xiong and
                  Yunchang Yang and
                  Di He and
                  Kai Zheng and
                  Shuxin Zheng and
                  Chen Xing and
                  Huishuai Zhang and
                  Yanyan Lan and
                  Liwei Wang and
                  Tie{-}Yan Liu},
  title        = {On Layer Normalization in the Transformer Architecture},
  booktitle    = {Proceedings of the 37th International Conference on Machine Learning,
                  {ICML} 2020, 13-18 July 2020, Virtual Event},
  series       = {Proceedings of Machine Learning Research},
  pages        = {10524--10533},
  publisher    = {{PMLR}},
  year         = {2020},
  url          = {http://proceedings.mlr.press/v119/xiong20b.html},
  bibsource    = {dblp computer science bibliography, https://dblp.org}
}

@misc{gelu,
      title={Gaussian Error Linear Units (GELUs)},
      author={Dan Hendrycks and Kevin Gimpel},
      year={2016},
      eprint={1606.08415},
      archivePrefix={arXiv},
      primaryClass={cs.LG},
      url={https://arxiv.org/abs/1606.08415},
}

@misc{swiglu,
      title={GLU Variants Improve Transformer},
      author={Noam Shazeer},
      year={2020},
      eprint={2002.05202},
      archivePrefix={arXiv},
      primaryClass={cs.LG},
      url={https://arxiv.org/abs/2002.05202},
}

@article{rope,
      title={RoFormer: Enhanced transformer with Rotary Position Embedding},
      author={Jianlin Su and Murtadha Ahmed and Yu Lu and Shengfeng Pan and Wen Bo and Yunfeng Liu},
      journal={Neurocomputing},
      volume={568},
      pages={127063},
      year={2024},
      doi={10.1016/j.neucom.2023.127063},
      url={https://doi.org/10.1016/j.neucom.2023.127063},
}

@misc{llama,
      title={LLaMA: Open and Efficient Foundation Language Models},
      author={Hugo Touvron and Thibaut Lavril and Gautier Izacard and Xavier Martinet and Marie-Anne Lachaux and Timothée Lacroix and Baptiste Rozière and Naman Goyal and Eric Hambro and Faisal Azhar and Aurelien Rodriguez and Armand Joulin and Edouard Grave and Guillaume Lample},
      year={2023},
      eprint={2302.13971},
      archivePrefix={arXiv},
      primaryClass={cs.CL},
      url={https://arxiv.org/abs/2302.13971},
}

@inproceedings{adamw,
      title = {Decoupled Weight Decay Regularization},
      author = {Ilya Loshchilov and Frank Hutter},
      year = {2019},
      url = {https://openreview.net/forum?id=Bkg6RiCqY7},
      booktitle = {7th International Conference on Learning Representations, ICLR 2019, New Orleans, LA, USA, May 6-9, 2019},
      publisher = {OpenReview.net},
}

@misc{muon,
  author       = {Keller Jordan and Yuchen Jin and Vlado Boza and Jiacheng You and
                  Franz Cesista and Laker Newhouse and Jeremy Bernstein},
  title        = {Muon: An optimizer for hidden layers in neural networks},
  year         = {2024},
  url          = {https://kellerjordan.github.io/posts/muon/}
}

@misc{moonshot_muon,
      title={Muon is Scalable for {LLM} Training},
      author={Jingyuan Liu and Jianlin Su and Xingcheng Yao and Zhejun Jiang and Guokun Lai and Yulun Du and Yidao Qin and Weixin Xu and Enzhe Lu and Junjie Yan and Yanru Chen and Huabin Zheng and Yibo Liu and Shaowei Liu and Bohong Yin and Weiran He and Han Zhu and Yuzhi Wang and Jianzhou Wang and Mengnan Dong and Zheng Zhang and Yongsheng Kang and Hao Zhang and Xinran Xu and Yutao Zhang and Yuxin Wu and Xinyu Zhou and Zhilin Yang},
      year={2025},
      eprint={2502.16982},
      archivePrefix={arXiv},
      primaryClass={cs.LG},
      url={https://arxiv.org/abs/2502.16982},
}

@article{efficient_inference,
  author       = {Reiner Pope and
                  Sholto Douglas and
                  Aakanksha Chowdhery and
                  Jacob Devlin and
                  James Bradbury and
                  Anselm Levskaya and
                  Jonathan Heek and
                  Kefan Xiao and
                  Shivani Agrawal and
                  Jeff Dean},
  title        = {Efficiently Scaling Transformer Inference},
  journal      = {CoRR},
  volume       = {abs/2211.05102},
  year         = {2022},
  url          = {https://doi.org/10.48550/arXiv.2211.05102},
  doi          = {10.48550/ARXIV.2211.05102},
  eprinttype   = {arXiv},
  eprint       = {2211.05102},
  bibsource    = {dblp computer science bibliography, https://dblp.org}
}

@article{syntaxofmatter,
author = {Anton Morgunov  and Yu Shee  and Alexander V Soudackov  and Victor S Batista },
title = {The Syntax of Matter: Synthesis Planning as the Foundation of Generative Chemistry},
journal = {ChemRxiv},
volume = {2026},
number = {0421},
pages = {},
year = {2026},
doi = {10.26434/chemrxiv.15001278/v3},
URL = {https://chemrxiv.org/doi/abs/10.26434/chemrxiv.15001278/v3},
eprint = {https://chemrxiv.org/doi/pdf/10.26434/chemrxiv.15001278/v3},
}

@Article{sascore_2009,
author={Ertl, Peter
and Schuffenhauer, Ansgar},
title={Estimation of synthetic accessibility score of drug-like molecules based on molecular complexity and fragment contributions},
journal={Journal of Cheminformatics},
year={2009},
day={10},
volume={1},
number={1},
pages={8},
issn={1758-2946},
doi={10.1186/1758-2946-1-8},
url={https://doi.org/10.1186/1758-2946-1-8}
}

@article{scscore_2018,
author = {Coley, Connor W. and Rogers, Luke and Green, William H. and Jensen, Klavs F.},
title = {SCScore: Synthetic Complexity Learned from a Reaction Corpus},
journal = {Journal of Chemical Information and Modeling},
volume = {58},
number = {2},
pages = {252-261},
year = {2018},
doi = {10.1021/acs.jcim.7b00622},
note ={PMID: 29309147},
URL = {https://doi.org/10.1021/acs.jcim.7b00622},
eprint = {https://doi.org/10.1021/acs.jcim.7b00622}
}

@article{rascore_2020,
  title={Retrosynthetic accessibility score ({RAscore}) -- rapid machine learned synthesizability classification from {AI} driven retrosynthetic planning},
  author={Amol Thakkar and Veronika Chadimov{\'a} and Esben Jannik Bjerrum and Ola Engkvist and Jean-Louis Reymond},
  journal={Chemical Science},
  year={2020},
  volume={12},
  pages={3339 - 3349},
  url={https://api.semanticscholar.org/CorpusID:233621461}
}

@Article{syba,
author={Vor{\v{s}}il{\'a}k, Milan and Kol{\'a}\v{r}, Micha{\l} and {\v{C}}melo, Ivan and Svozil, Daniel},
title={SYBA: Bayesian estimation of synthetic accessibility of organic compounds},
journal={Journal of Cheminformatics},
year={2020},
volume={12},
number={1},
pages={35},
issn={1758-2946},
doi={10.1186/s13321-020-00439-2},
url={https://doi.org/10.1186/s13321-020-00439-2}
}

@article{corey_1969,
  title = {Computer-Assisted Design of Complex Organic Syntheses: Pathways for molecular synthesis can be devised with a computer and equipment for graphical communication.},
  volume = {166},
  issn = {0036-8075, 1095-9203},
  url = {https://www.science.org/doi/10.1126/science.166.3902.178},
  doi = {10.1126/science.166.3902.178},
  number = {3902},
  journal = {Science},
  author = {Corey, E. J. and Wipke, W. Todd},
  year = {1969},
  pages = {178--192}
}

@Article{mcts_2018,
author={Segler, Marwin H. S.
and Preuss, Mike
and Waller, Mark P.},
title={Planning chemical syntheses with deep neural networks and symbolic {AI}},
journal={Nature},
year={2018},
volume={555},
number={7698},
pages={604-610},
issn={1476-4687},
doi={10.1038/nature25978},
url={https://doi.org/10.1038/nature25978}
}

@inproceedings{retrostar_2020,
      title={{Retro*}: Learning Retrosynthetic Planning with Neural Guided {A*} Search},
      author={Binghong Chen and Chengtao Li and Hanjun Dai and Le Song},
      booktitle={Proceedings of the 37th International Conference on Machine Learning},
      series={Proceedings of Machine Learning Research},
      volume={119},
      pages={1608--1616},
      publisher={PMLR},
      year={2020},
      url={https://proceedings.mlr.press/v119/chen20k.html},
}

@Article{aizyn_2020,
author={Genheden, Samuel
and Thakkar, Amol
and Chadimov{\'a}, Veronika
and Reymond, Jean-Louis
and Engkvist, Ola
and Bjerrum, Esben},
title={{AiZynthFinder}: a fast, robust and flexible open-source software for retrosynthetic planning},
journal={Journal of Cheminformatics},
year={2020},
volume={12},
number={1},
pages={70},
issn={1758-2946},
doi={10.1186/s13321-020-00472-1},
url={https://doi.org/10.1186/s13321-020-00472-1}
}

@article{moltransformer_2020,
  title={Predicting retrosynthetic pathways using transformer-based models and a hyper-graph exploration strategy},
  author={Philippe Schwaller and Riccardo Petraglia and Valerio Zullo and Vishnu H. Nair and Rico H{\"a}uselmann and Riccardo Pisoni and Costas Bekas and Anna Iuliano and Teodoro Laino},
  journal={Chemical Science},
  year={2020},
  volume={11},
  pages={3316 - 3325},
  url={https://api.semanticscholar.org/CorpusID:216332642}
}

@inproceedings{dfpn_2019,
 author = {Kishimoto, Akihiro and Buesser, Beat and Chen, Bei and Botea, Adi},
 booktitle = {Advances in Neural Information Processing Systems},
 editor = {H. Wallach and H. Larochelle and A. Beygelzimer and F. d\textquotesingle Alch{\'e}-Buc and E. Fox and R. Garnett},
 pages = {},
 publisher = {Curran Associates, Inc.},
 title = {Depth-First Proof-Number Search with Heuristic Edge Cost and Application to Chemical Synthesis Planning},
 url = {https://proceedings.neurips.cc/paper_files/paper/2019/file/4fc28b7093b135c21c7183ac07e928a6-Paper.pdf},
 volume = {32},
 year = {2019}
}

@inproceedings{grasp_2022,
 author = {Yu, Yemin and Wei, Ying and Kuang, Kun and Huang, Zhengxing and Yao, Huaxiu and Wu, Fei},
 booktitle = {Advances in Neural Information Processing Systems},
 editor = {S. Koyejo and S. Mohamed and A. Agarwal and D. Belgrave and K. Cho and A. Oh},
 pages = {10257--10268},
 publisher = {Curran Associates, Inc.},
 title = {GRASP: Navigating Retrosynthetic Planning with Goal-driven Policy},
 url = {https://proceedings.neurips.cc/paper_files/paper/2022/file/42beaab8aa8da1c77581609a61eced93-Paper-Conference.pdf},
 volume = {35},
 year = {2022}
}

@inproceedings{retrograph_2022,
author = {Xie, Shufang and Yan, Rui and Han, Peng and Xia, Yingce and Wu, Lijun and Guo, Chenjuan and Yang, Bin and Qin, Tao},
title = {RetroGraph: Retrosynthetic Planning with Graph Search},
year = {2022},
isbn = {9781450393850},
publisher = {Association for Computing Machinery},
address = {New York, NY, USA},
url = {https://doi.org/10.1145/3534678.3539446},
doi = {10.1145/3534678.3539446},
booktitle = {Proceedings of the 28th ACM SIGKDD Conference on Knowledge Discovery and Data Mining},
pages = {2120-2129},
numpages = {10},
location = {Washington DC, USA},
series = {KDD '22}
}

@Article{egmcts_2023,
author={Hong, Siqi and Zhuo, Hankz Hankui and Jin, Kebing and Shao, Guang and Zhou, Zhanwen},
title={Retrosynthetic planning with experience-guided Monte Carlo tree search},
journal={Communications Chemistry},
year={2023},
volume={6},
number={1},
pages={120},
doi={10.1038/s42004-023-00911-8},
url={https://doi.org/10.1038/s42004-023-00911-8},
}

@article{meea_2024,
  title={Efficient retrosynthetic planning with {MCTS} exploration enhanced {A*} search},
  author={Dengwei Zhao and Shikui Tu and Lei Xu},
  journal={Communications Chemistry},
  year={2024},
  volume={7},
  url={https://api.semanticscholar.org/CorpusID:268252759}
}

@article{higherlev_2025,
  title={Higher-Level Strategies for Computer-Aided Retrosynthesis},
  author={Roh, Jihye and Joung, Joonyoung F. and Yu, Kevin and Tu, Zhengkai and Bartholomew, G. Logan and Santiago-Reyes, Omar A. and Fong, Mun Hong and Sarpong, Richmond and Reisman, Sarah E. and Coley, Connor W.},
  journal={ACS Central Science},
  volume={12},
  number={3},
  pages={345--357},
  year={2026},
  doi={10.1021/acscentsci.5c02014},
  url={https://doi.org/10.1021/acscentsci.5c02014},
}

@article{synplanner_2025,
author = {Akhmetshin, Tagir and Zankov, Dmitry and Gantzer, Philippe and Babadeev, Dmitry and Pinigina, Anna and Madzhidov, Timur and Varnek, Alexandre},
title = {SynPlanner: An End-to-End Tool for Synthesis Planning},
journal = {Journal of Chemical Information and Modeling},
volume = {65},
number = {1},
pages = {15-21},
year = {2025},
doi = {10.1021/acs.jcim.4c02004},
}

@misc{treemdp_2025,
      title={Retrosynthesis Planning via Worst-path Policy Optimisation in Tree-structured MDPs}, 
      author={Mianchu Wang and Giovanni Montana},
      year={2025},
      eprint={2509.10504},
      archivePrefix={arXiv},
      primaryClass={cs.LG},
      url={https://arxiv.org/abs/2509.10504}, 
}

@article{autosynroute_2020,
  title={Automatic retrosynthetic route planning using template-free models},
  author={Kangjie Lin and Youjun Xu and Jianfeng Pei and Luhua Lai},
  journal={Chemical Science},
  year={2020},
  volume={11},
  pages={3355 - 3364},
  url={https://api.semanticscholar.org/CorpusID:268816571}
}

@article{synllama_2025,
author = {Sun, Kunyang and Bagni, Dorian and Cavanagh, Joseph M. and Wang, Yingze and Sawyer, Jacob M. and Zhou, Bo and Gritsevskiy, Andrew and Zhang, Oufan and Head-Gordon, Teresa},
title = {SynLlama: Generating Synthesizable Molecules and Their Analogs with Large Language Models},
journal = {ACS Central Science},
volume = {11},
number = {11},
pages = {2108-2120},
year = {2025},
doi = {10.1021/acscentsci.5c01285},
}

@Article{retrosynformer_2026,
author ="Granqvist, Emma and Mercado, Roc{\'i}o and Genheden, Samuel",
title  ="Retrosynformer: planning multi-step chemical synthesis routes via a decision transformer",
journal  ="Digital Discovery",
year  ="2026",
volume  ="5",
issue  ="1",
pages  ="348-362",
publisher  ="RSC",
doi  ="10.1039/D5DD00153F",
url  ="http://dx.doi.org/10.1039/D5DD00153F",
}

@misc{llamole_2024,
      title={Multimodal Large Language Models for Inverse Molecular Design with Retrosynthetic Planning}, 
      author={Gang Liu and Michael Sun and Wojciech Matusik and Meng Jiang and Jie Chen},
      year={2024},
      eprint={2410.04223},
      archivePrefix={arXiv},
      primaryClass={cs.LG},
      url={https://arxiv.org/abs/2410.04223}, 
}

@inproceedings{desp_2024,
      title={Double-Ended Synthesis Planning with Goal-Constrained Bidirectional Search},
      author={Kevin Yu and Jihye Roh and Ziang Li and Wenhao Gao and Runzhong Wang and Connor W. Coley},
      booktitle={Advances in Neural Information Processing Systems},
      volume={37},
      pages={112919--112949},
      publisher={Neural Information Processing Systems Foundation, Inc.},
      year={2024},
      doi={10.52202/079017-3588},
      url={https://proceedings.neurips.cc/paper_files/paper/2024/hash/cd091a4d8e97157d32940428f902c7b0-Abstract-Conference.html},
}

@misc{chimera_2025,
      title={Chemist-aligned retrosynthesis by ensembling diverse inductive bias models}, 
      author={Krzysztof Maziarz and Guoqing Liu and Hubert Misztela and Austin Tripp and Junren Li and Aleksei Kornev and Piotr Gai{\'n}ski and Holger Hoefling and Mike Fortunato and Rishi Gupta and Marwin Segler},
      year={2025},
      eprint={2412.05269},
      archivePrefix={arXiv},
      primaryClass={cs.LG},
      url={https://arxiv.org/abs/2412.05269}, 
}

@inproceedings{larc_2025,
      title={{LARC}: Towards Human-level Constrained Retrosynthesis Planning through an Agentic Framework},
      author={Frazier N. Baker and Daniel Adu-Ampratwum and Reza Averly and Botao Yu and Huan Sun and Xia Ning},
      booktitle={Proceedings of AI for Accelerated Research Symposium},
      series={EPiC Series in Technology},
      volume={3},
      pages={153--176},
      publisher={EasyChair},
      year={2026},
      doi={10.29007/z3hb},
      url={https://easychair.org/publications/paper/SMVW},
}

@misc{synthelite_2025,
      title={Synthelite: Chemist-aligned and feasibility-aware synthesis planning with {LLMs}},
      author={Nguyen Xuan-Vu and Daniel Armstrong and Milena Wehrbach and Andres M. Bran and Zlatko Jon{\v{c}}ev and Philippe Schwaller},
      year={2025},
      eprint={2512.16424},
      archivePrefix={arXiv},
      primaryClass={cs.AI},
      url={https://arxiv.org/abs/2512.16424},
}

@misc{retrofallback_2024,
      title={Retro-fallback: retrosynthetic planning in an uncertain world},
      author={Austin Tripp and Krzysztof Maziarz and Sarah Lewis and Marwin Segler and Jos{\'e} Miguel Hern{\'a}ndez-Lobato},
      year={2024},
      eprint={2310.09270},
      archivePrefix={arXiv},
      primaryClass={cs.AI},
      url={https://arxiv.org/abs/2310.09270},
}

@misc{song_2025,
      title={{AOT*}: Efficient Synthesis Planning via {LLM}-Empowered {AND-OR} Tree Search}, 
      author={Xiaozhuang Song and Xuanhao Pan and Xinjian Zhao and Hangting Ye and Shufei Zhang and Jian Tang and Tianshu Yu},
      year={2025},
      eprint={2509.20988},
      archivePrefix={arXiv},
      primaryClass={cs.AI},
      url={https://arxiv.org/abs/2509.20988}, 
}

@misc{tempre_2025,
      title={{TempRe}: Template generation for single and direct multi-step retrosynthesis}, 
      author={Nguyen Xuan-Vu and Daniel P Armstrong and Zlatko Jon{\v{c}}ev and Philippe Schwaller},
      year={2025},
      eprint={2507.21762},
      archivePrefix={arXiv},
      primaryClass={cs.LG},
      url={https://arxiv.org/abs/2507.21762}, 
}

@misc{llmmulti_2025,
      title={{LLM}-Augmented Chemical Synthesis and Design Decision Programs}, 
      author={Haorui Wang and Jeff Guo and Lingkai Kong and Rampi Ramprasad and Philippe Schwaller and Yuanqi Du and Chao Zhang},
      year={2025},
      eprint={2505.07027},
      archivePrefix={arXiv},
      primaryClass={cs.AI},
      url={https://arxiv.org/abs/2505.07027}, 
}

@inproceedings{liu_2023,
  title={Retrosynthetic Planning with Dual Value Networks},
  author={Guoqing Liu and Di Xue and Shufang Xie and Yingce Xia and Austin Tripp and Krzysztof Maziarz and Marwin H. S. Segler and Tao Qin and Zongzhang Zhang and Tie-Yan Liu},
  booktitle={International Conference on Machine Learning},
  year={2023},
  url={https://api.semanticscholar.org/CorpusID:256416110}
}

@article{enhmcts_2025,
  title={Enhancing {Monte Carlo Tree Search} for Retrosynthesis},
  author={Ton M Blackshaw and Joseph C. Davies and Kristian T Spoerer and Jonathan D. Hirst},
  journal={Journal of Chemical Information and Modeling},
  year={2025},
  volume={65},
  pages={6537 - 6546},
  url={https://api.semanticscholar.org/CorpusID:279328860}
}

@article{resynz_2024,
  title={Retrosynthesis Zero: Self-Improving Global Synthesis Planning Using Reinforcement Learning.},
  author={Jiasheng Guo and Chenning Yu and Kenan Li and Yijian Zhang and Guoqiang Wang and Shuhua Li and Hao Dong},
  journal={Journal of chemical theory and computation},
  year={2024},
  url={https://api.semanticscholar.org/CorpusID:269771006}
}

@Article{green_mcts_2020,
author ="Wang, Xiaoxue and Qian, Yujie and Gao, Hanyu and Coley, Connor W. and Mo, Yiming and Barzilay, Regina and Jensen, Klavs F.",
title  ="Towards efficient discovery of green synthetic pathways with Monte Carlo tree search and reinforcement learning",
journal  ="Chem. Sci.",
year  ="2020",
volume  ="11",
issue  ="40",
pages  ="10959-10972",
publisher  ="The Royal Society of Chemistry",
doi  ="10.1039/D0SC04184J",
url  ="http://dx.doi.org/10.1039/D0SC04184J",
}

@article{selfplay_2019,
author = {Schreck, John S. and Coley, Connor W. and Bishop, Kyle J. M.},
title = {Learning Retrosynthetic Planning through Simulated Experience},
journal = {ACS Central Science},
volume = {5},
number = {6},
pages = {970-981},
year = {2019},
doi = {10.1021/acscentsci.9b00055},
}

@article{retrek_2022,
author = {Ishida, Shoichi and Terayama, Kei and Kojima, Ryosuke and Takasu, Kiyosei and Okuno, Yasushi},
title = {{AI}-Driven Synthetic Route Design Incorporated with Retrosynthesis Knowledge},
journal = {Journal of Chemical Information and Modeling},
volume = {62},
number = {6},
pages = {1357-1367},
year = {2022},
doi = {10.1021/acs.jcim.1c01074},
}

@article{nested_mcts_2024,
  title={Comparing search algorithms on the retrosynthesis problem},
  author={Milo Roucairol and Tristan Cazenave},
  journal={Molecular Informatics},
  year={2024},
  volume={43},
  url={https://api.semanticscholar.org/CorpusID:253882602}
}

@article{multistepttl_2023,
  title={Multistep retrosynthesis combining a disconnection aware triple transformer loop with a route penalty score guided tree search},
  author={David Kreutter and Jean-Louis Reymond},
  journal={Chemical Science},
  year={2023},
  volume={14},
  pages={9959 - 9969},
  url={https://api.semanticscholar.org/CorpusID:261488982}
}

@article{evoretro_2023,
  title={Evolutionary Retrosynthetic Route Planning [Research Frontier]},
  author={Yan Zhang and Xiao He and Shuanhu Gao and Aimin Zhou and Hao Hao},
  journal={IEEE Computational Intelligence Magazine},
  year={2023},
  volume={19},
  pages={58-72},
  url={https://api.semanticscholar.org/CorpusID:271115363}
}

@Article{dreamretroer_2025,
author={Zhang, Xuefeng and Lin, Haowei and Zhang, Muhan and Zhou, Yuan and Ma, Jianzhu},
title={A data-driven group retrosynthesis planning model inspired by neurosymbolic programming},
journal={Nature Communications},
year={2025},
volume={16},
number={1},
pages={192},
doi={10.1038/s41467-024-55374-9},
url={https://doi.org/10.1038/s41467-024-55374-9},
}

@article{retrogfn_2025,
      title={Diverse and feasible retrosynthesis using {GFlowNets}},
      author={Piotr Gai{\'n}ski and Micha\l{} Koziarski and Krzysztof Maziarz and Marwin Segler and Jacek Tabor and Marek {\'S}mieja},
      journal={Information Sciences},
      volume={714},
      pages={122194},
      year={2025},
      doi={10.1016/j.ins.2025.122194},
      url={https://doi.org/10.1016/j.ins.2025.122194},
}

@Article{andronov_2025,
      title={Fast and scalable retrosynthetic planning with a transformer neural network and speculative beam search},
      author={Natalia Andronova and Mikhail Andronov and J{\"u}rgen Schmidhuber and Michael Wand and Djork-Arn{\'e} Clevert},
      journal={Digital Discovery},
      volume={5},
      issue={4},
      pages={1783--1793},
      year={2026},
      publisher={RSC},
      doi={10.1039/D5DD00573F},
      url={http://dx.doi.org/10.1039/D5DD00573F},
}

@misc{bitter_lesson,
      title={The Bitter Lesson},
      author={Sutton, Richard S.},
      year={2019},
      url={http://www.incompleteideas.net/IncIdeas/BitterLesson.html},
}

@Article{paroutes,
author ="Genheden, Samuel and Bjerrum, Esben",
title  ="{PaRoutes}: towards a framework for benchmarking retrosynthesis route predictions",
journal  ="Digital Discovery",
year  ="2022",
volume  ="1",
issue  ="4",
pages  ="527-539",
publisher  ="RSC",
doi  ="10.1039/D2DD00015F",
url  ="http://dx.doi.org/10.1039/D2DD00015F",
}

@article{parrot_2024,
  title={Integrating synthetic accessibility with {AI}-based generative drug design},
  author={Maud Parrot and Hamza Tajmouati and Vinicius Barros Ribeiro da Silva and Brian Ross Atwood and Robin Fourcade and Yann Gaston-Math{\'e} and Nicolas Do Huu and Quentin Perron},
  journal={Journal of Cheminformatics},
  year={2021},
  volume={15},
  url={https://api.semanticscholar.org/CorpusID:245417923}
}

@article{synthformer_2025,
      title={SynthFormer: Equivariant pharmacophore-based generation of synthesizable molecules for ligand-based drug design},
      author={Zygimantas Jocys and Zhanxing Zhu and Henriette M. G. Willems and Katayoun Farrahi},
      journal={Artificial Intelligence in the Life Sciences},
      volume={9},
      pages={100148},
      year={2026},
      doi={10.1016/j.ailsci.2025.100148},
      url={https://doi.org/10.1016/j.ailsci.2025.100148},
}

@misc{rxnflow_2025,
      title={Generative Flows on Synthetic Pathway for Drug Design}, 
      author={Seonghwan Seo and Minsu Kim and Tony Shen and Martin Ester and Jinkyoo Park and Sungsoo Ahn and Woo Youn Kim},
      year={2025},
      eprint={2410.04542},
      archivePrefix={arXiv},
      primaryClass={q-bio.BM},
      url={https://arxiv.org/abs/2410.04542}, 
}

@inproceedings{rgfn_2024,
      title={{RGFN}: Synthesizable Molecular Generation Using {GFlowNets}},
      author={Micha\l{} Koziarski and Andrei Rekesh and Dmytro Shevchuk and Almer van der Sloot and Piotr Gai{\'n}ski and Yoshua Bengio and Cheng-Hao Liu and Mike Tyers and Robert A. Batey},
      booktitle={Advances in Neural Information Processing Systems},
      volume={37},
      pages={46908--46955},
      publisher={Neural Information Processing Systems Foundation, Inc.},
      year={2024},
      doi={10.52202/079017-1488},
      url={https://proceedings.neurips.cc/paper_files/paper/2024/hash/53704142f230054140418ecd8857f391-Abstract-Conference.html},
}

@article{grandchallenges,
author = {Connor W Coley  and Pankaj Daga  and Marco De Vivo  and Willem Jespers  and Ashutosh S Jogalekar  and S Roy Kimura  and Lucien Koenekoop  and Anne-Grete Märtson  and Timothy R Newhouse  and Soumya Ray  and Riccardo Sabatini  and David C Thompson  and Woody Sherman },
title = {Grand Challenges for Predictive Modeling in Small Molecule Drug Discovery},
journal = {ChemRxiv},
volume = {2026},
number = {0304},
pages = {},
year = {2026},
doi = {10.26434/chemrxiv.15000615/v1},
URL = {https://chemrxiv.org/doi/abs/10.26434/chemrxiv.15000615/v1},
eprint = {https://chemrxiv.org/doi/pdf/10.26434/chemrxiv.15000615/v1},
}

@misc{prexsyn_2025,
      title={Efficient and Programmable Exploration of Synthesizable Chemical Space}, 
      author={Shitong Luo and Connor W. Coley},
      year={2025},
      eprint={2512.00384},
      archivePrefix={arXiv},
      primaryClass={cs.LG},
      url={https://arxiv.org/abs/2512.00384}, 
}

@article{askcos_2025,
author = {Tu, Zhengkai and Choure, Sourabh J. and Fong, Mun Hong and Roh, Jihye and Levin, Itai and Yu, Kevin and Joung, Joonyoung F. and Morgan, Nathan and Li, Shih-Cheng and Sun, Xiaoqi and Lin, Huiqian and Murnin, Mark and Liles, Jordan P. and Struble, Thomas J. and Fortunato, Michael E. and Liu, Mengjie and Green, William H. and Jensen, Klavs F. and Coley, Connor W.},
title = {{ASKCOS}: Open-Source, Data-Driven Synthesis Planning},
journal = {Accounts of Chemical Research},
volume = {58},
number = {11},
pages = {1764-1775},
year = {2025},
doi = {10.1021/acs.accounts.5c00155},
}

@inproceedings{prefixlm_2019,
      title={Unified Language Model Pre-training for Natural Language Understanding and Generation},
      author={Li Dong and Nan Yang and Wenhui Wang and Furu Wei and Xiaodong Liu and Yu Wang and Jianfeng Gao and Ming Zhou and Hsiao-Wuen Hon},
      booktitle={Advances in Neural Information Processing Systems},
      volume={32},
      year={2019},
      url={https://proceedings.neurips.cc/paper/2019/hash/c20bb2d9a50d5ac1f713f8b34d9aac5a-Abstract.html},
}
}

\clearpage
\newpage
\renewcommand{\thetable}{S\arabic{table}}
\renewcommand{\thefigure}{S\arabic{figure}}
\setcounter{table}{0}
\setcounter{figure}{0}

\section*{Supporting Information}

The Supporting Information reports additional target-only evaluations that contextualize the main \texttt{mkt-cnv-160} results.
Tables~\ref{tab:si-mkt-lin-500-results}--\ref{tab:si-ref-lng-84-results} extend the Ariadne versus AiZynthFinder MCTS comparison to \texttt{mkt-lin-500}, \texttt{ref-cnv-400}, \texttt{ref-lin-600}, and \texttt{ref-lng-84}, with paired confidence intervals for Solv-0, Top-1, Top-10, root recovery, and route recovery conditional on the root.
Table~\ref{tab:si-mkt-lin-500-beam-sweep} reports the \texttt{mkt-lin-500} beam-size sweep for the 24-layer Ariadne checkpoint.

\paragraph{Supplementary Tables S1--S4}
These tables report the full target-only benchmark suite beyond \texttt{mkt-cnv-160}.
Table~\ref{tab:si-mkt-lin-500-results} evaluates the commercial-stock linear benchmark.
Tables~\ref{tab:si-ref-cnv-400-results}, \ref{tab:si-ref-lin-600-results}, and \ref{tab:si-ref-lng-84-results} evaluate the reference-stock convergent, linear, and long-route benchmarks.

\paragraph{Supplementary Table S5}
Table~\ref{tab:si-mkt-lin-500-beam-sweep} reports the effect of beam size on \texttt{mkt-lin-500} for the 24-layer Ariadne checkpoint.

\begin{table}[H]
\centering
\scriptsize
\resizebox{\textwidth}{!}{%
\begin{tabular}{l r r r r r r r r}
\toprule
Model & Tier-0 valid & Solv-0 & Top-1 & Top-10 & Top-10 CI & Root T10 & Route$\mid$root T10 & Time \\
\midrule
\multicolumn{9}{l}{\textit{Training on \texttt{v2026-05-12-reaction}}} \\
AiZynthFinder MCTS & 100.0\% & \b{92.4\%} & 17.2\% & 30.6\% & [26.6, 34.6] & 56.6\% & 54.1\% & 48.4 min \\
Ariadne 24L & 100.0\% & 90.4\% & \b{18.8\%} & \b{37.2\%} & [33.0, 41.4] & 63.0\% & 59.0\% & 1.0 hr \\
\multicolumn{2}{l}{\textit{Paired 95\% CI (Ariadne--MCTS)}} & [-5.2, +1.4] & [-1.8, +5.0] & [+2.2, +10.8] & -- & [+1.2, +11.6] & [-1.0, +10.9] & -- \\
\addlinespace
\multicolumn{9}{l}{\textit{Training on \texttt{v2026-05-12-route}}} \\
AiZynthFinder MCTS & 100.0\% & 92.6\% & 18.6\% & 33.6\% & [29.4, 37.8] & 58.8\% & 57.1\% & 48.1 min \\
Ariadne 24L & 100.0\% & \b{96.4\%} & \b{29.4\%} & \b{52.6\%} & [48.2, 57.0] & 72.4\% & 72.7\% & 1.1 hr \\
\multicolumn{2}{l}{\textit{Paired 95\% CI (Ariadne--MCTS)}} & [+1.4, +6.2] & [+6.4, +15.2] & [+14.0, +23.8] & -- & [+8.8, +18.4] & [+9.5, +21.6] & -- \\
\bottomrule
\end{tabular}}
\caption{Additional \texttt{mkt-lin-500} results. Ariadne uses the 24-layer checkpoint with target-only prompting and beam size 50. AiZynthFinder uses 100 MCTS iterations and maximum search depth 10. Top-10 CI gives the 95\% bootstrap interval. Root T10 measures recovery of the reference root reaction, and Route$\mid$root T10 measures full-route recovery among targets whose root reaction was recovered. Paired CI rows compare Ariadne with AiZynthFinder MCTS: intervals containing zero indicate no significant difference, positive intervals favor Ariadne, and negative intervals favor AiZynthFinder MCTS.}
\label{tab:si-mkt-lin-500-results}
\end{table}

\begin{table}[H]
\centering
\scriptsize
\resizebox{\textwidth}{!}{%
\begin{tabular}{l r r r r r r r r}
\toprule
Model & Tier-0 valid & Solv-0 & Top-1 & Top-10 & Top-10 CI & Root T10 & Route$\mid$root T10 & Time \\
\midrule
\multicolumn{9}{l}{\textit{Training on \texttt{v2026-05-12-reaction}}} \\
AiZynthFinder MCTS & 100.0\% & \b{76.5\%} & \b{16.8\%} & \b{24.5\%} & [20.2, 28.7] & 54.8\% & 44.7\% & 1.3 hr \\
Ariadne 24L & 100.0\% & 33.0\% & 8.0\% & 9.8\% & [7.0, 12.8] & 18.0\% & 54.2\% & 1.0 hr \\
\multicolumn{2}{l}{\textit{Paired 95\% CI (Ariadne--MCTS)}} & [-49.0, -37.8] & [-12.5, -5.0] & [-19.0, -10.5] & -- & [-42.2, -31.5] & [-2.8, +21.7] & -- \\
\addlinespace
\multicolumn{9}{l}{\textit{Training on \texttt{v2026-05-12-route}}} \\
AiZynthFinder MCTS & 100.0\% & \b{85.2\%} & 23.8\% & 36.2\% & [31.5, 41.0] & 63.2\% & 57.3\% & 1.1 hr \\
Ariadne 24L & 100.0\% & 68.2\% & \b{30.5\%} & \b{41.5\%} & [36.8, 46.2] & 49.8\% & 83.4\% & 1.1 hr \\
\multicolumn{2}{l}{\textit{Paired 95\% CI (Ariadne--MCTS)}} & [-22.2, -12.0] & [+2.0, +11.5] & [-0.3, +10.7] & -- & [-19.5, -7.8] & [+18.8, +33.3] & -- \\
\bottomrule
\end{tabular}}
\caption{Additional \texttt{ref-cnv-400} results. Ariadne uses the 24-layer checkpoint with target-only prompting and beam size 50. AiZynthFinder uses 100 MCTS iterations and maximum search depth 10. Top-10 CI gives the 95\% bootstrap interval. Root T10 measures recovery of the reference root reaction, and Route$\mid$root T10 measures full-route recovery among targets whose root reaction was recovered. Paired CI rows compare Ariadne with AiZynthFinder MCTS: intervals containing zero indicate no significant difference, positive intervals favor Ariadne, and negative intervals favor AiZynthFinder MCTS.}
\label{tab:si-ref-cnv-400-results}
\end{table}

\begin{table}[H]
\centering
\scriptsize
\resizebox{\textwidth}{!}{%
\begin{tabular}{l r r r r r r r r}
\toprule
Model & Tier-0 valid & Solv-0 & Top-1 & Top-10 & Top-10 CI & Root T10 & Route$\mid$root T10 & Time \\
\midrule
\multicolumn{9}{l}{\textit{Training on \texttt{v2026-05-12-reaction}}} \\
AiZynthFinder MCTS & 100.0\% & \b{78.8\%} & 10.5\% & 19.5\% & [16.3, 22.7] & 47.2\% & 41.3\% & 1.5 hr \\
Ariadne 24L & 100.0\% & 55.8\% & \b{13.2\%} & \b{21.0\%} & [17.7, 24.3] & 36.8\% & 57.0\% & 1.5 hr \\
\multicolumn{2}{l}{\textit{Paired 95\% CI (Ariadne--MCTS)}} & [-27.5, -18.5] & [-0.3, +5.7] & [-2.3, +5.3] & -- & [-15.5, -5.2] & [+7.8, +23.5] & -- \\
\addlinespace
\multicolumn{9}{l}{\textit{Training on \texttt{v2026-05-12-route}}} \\
AiZynthFinder MCTS & 100.0\% & \b{83.2\%} & 14.2\% & 26.3\% & [22.8, 29.8] & 49.7\% & 53.0\% & 1.5 hr \\
Ariadne 24L & 100.0\% & 74.8\% & \b{26.0\%} & \b{40.0\%} & [36.2, 44.0] & 55.3\% & 72.3\% & 1.6 hr \\
\multicolumn{2}{l}{\textit{Paired 95\% CI (Ariadne--MCTS)}} & [-12.3, -4.3] & [+7.8, +15.7] & [+9.3, +18.0] & -- & [+0.8, +10.5] & [+12.7, +25.7] & -- \\
\bottomrule
\end{tabular}}
\caption{Additional \texttt{ref-lin-600} results. Ariadne uses the 24-layer checkpoint with target-only prompting and beam size 50. AiZynthFinder uses 100 MCTS iterations and maximum search depth 10. Top-10 CI gives the 95\% bootstrap interval. Root T10 measures recovery of the reference root reaction, and Route$\mid$root T10 measures full-route recovery among targets whose root reaction was recovered. Paired CI rows compare Ariadne with AiZynthFinder MCTS: intervals containing zero indicate no significant difference, positive intervals favor Ariadne, and negative intervals favor AiZynthFinder MCTS.}
\label{tab:si-ref-lin-600-results}
\end{table}

\begin{table}[H]
\centering
\scriptsize
\resizebox{\textwidth}{!}{%
\begin{tabular}{l r r r r r r r r}
\toprule
Model & Tier-0 valid & Solv-0 & Top-1 & Top-10 & Top-10 CI & Root T10 & Route$\mid$root T10 & Time \\
\midrule
\multicolumn{9}{l}{\textit{Training on \texttt{v2026-05-12-reaction}}} \\
AiZynthFinder MCTS & 100.0\% & \b{59.5\%} & \b{0.0\%} & 0.0\% & [0.0, 0.0] & 27.4\% & 0.0\% & 15.0 min \\
Ariadne 24L & 100.0\% & 35.7\% & \b{0.0\%} & \b{2.4\%} & [0.0, 6.0] & 22.6\% & 10.5\% & 17.0 min \\
\multicolumn{2}{l}{\textit{Paired 95\% CI (Ariadne--MCTS)}} & [-38.1, -9.5] & [+0.0, +0.0] & [+0.0, +6.0] & -- & [-17.9, +7.1] & [+0.0, +26.7] & -- \\
\addlinespace
\multicolumn{9}{l}{\textit{Training on \texttt{v2026-05-12-route}}} \\
AiZynthFinder MCTS & 100.0\% & \b{70.2\%} & 0.0\% & 4.8\% & [1.2, 9.5] & 45.2\% & 10.5\% & 13.3 min \\
Ariadne 24L & 100.0\% & 66.7\% & \b{31.0\%} & \b{44.0\%} & [33.3, 54.8] & 53.6\% & 82.2\% & 21.0 min \\
\multicolumn{2}{l}{\textit{Paired 95\% CI (Ariadne--MCTS)}} & [-15.5, +8.3] & [+21.4, +40.5] & [+27.4, +51.2] & -- & [-3.6, +20.2] & [+56.5, +85.4] & -- \\
\bottomrule
\end{tabular}}
\caption{Additional \texttt{ref-lng-84} results. Ariadne uses the 24-layer checkpoint with target-only prompting and beam size 50. AiZynthFinder uses 100 MCTS iterations and maximum search depth 10. Top-10 CI gives the 95\% bootstrap interval. Root T10 measures recovery of the reference root reaction, and Route$\mid$root T10 measures full-route recovery among targets whose root reaction was recovered. Paired CI rows compare Ariadne with AiZynthFinder MCTS: intervals containing zero indicate no significant difference, positive intervals favor Ariadne, and negative intervals favor AiZynthFinder MCTS.}
\label{tab:si-ref-lng-84-results}
\end{table}

\begin{table}[H]
\centering
\scriptsize
\resizebox{\textwidth}{!}{%
\begin{tabular}{r r r r r r r r r}
\toprule
Beam size & Tier-0 valid & Solv-0 & Top-1 & Top-10 & Top-10 CI & Root T10 & Route$\mid$root T10 & Time \\
\midrule
\multicolumn{9}{l}{\textit{Training on \texttt{v2026-05-12-reaction}}} \\
1 & 98.2\% & 33.8\% & 10.0\% & 10.0\% & [7.4, 12.8] & 17.4\% & 57.5\% & 25.4 min \\
5 & 99.4\% & 57.6\% & 17.0\% & 21.0\% & [17.4, 24.8] & 32.6\% & 64.4\% & 37.4 min \\
10 & 99.8\% & 69.0\% & 18.2\% & 26.0\% & [22.2, 30.0] & 42.0\% & 61.9\% & 43.4 min \\
20 & 100.0\% & 78.4\% & \b{19.2\%} & 31.2\% & [27.2, 35.2] & 52.2\% & 59.8\% & 49.3 min \\
30 & 100.0\% & 85.2\% & 19.0\% & 34.0\% & [29.8, 38.0] & 58.0\% & 58.6\% & 54.3 min \\
50 & 100.0\% & \b{90.4\%} & 18.8\% & \b{37.2\%} & [33.0, 41.4] & 63.0\% & 59.0\% & 1.0 hr \\
\addlinespace
\multicolumn{9}{l}{\textit{Training on \texttt{v2026-05-12-route}}} \\
1 & 97.0\% & 42.2\% & 15.6\% & 15.6\% & [12.4, 18.8] & 22.0\% & 70.9\% & 30.2 min \\
5 & 99.6\% & 74.0\% & 26.4\% & 32.4\% & [28.4, 36.6] & 46.2\% & 70.1\% & 42.7 min \\
10 & 99.8\% & 82.0\% & 27.8\% & 38.0\% & [33.8, 42.2] & 53.6\% & 70.9\% & 49.0 min \\
20 & 100.0\% & 89.2\% & 29.2\% & 45.8\% & [41.4, 50.2] & 63.8\% & 71.8\% & 54.7 min \\
30 & 100.0\% & 93.4\% & \b{29.4\%} & 48.2\% & [43.8, 52.6] & 68.4\% & 70.5\% & 59.7 min \\
50 & 100.0\% & \b{96.4\%} & \b{29.4\%} & \b{52.6\%} & [48.2, 57.0] & 72.4\% & 72.7\% & 1.1 hr \\
\bottomrule
\end{tabular}}
\caption{\texttt{mkt-lin-500} beam-size sweep for the 24-layer Ariadne checkpoint with target-only prompting. Top-10 CI gives the 95\% bootstrap interval. Root T10 measures recovery of the reference root reaction, and Route$\mid$root T10 measures full-route recovery among targets whose root reaction was recovered.}
\label{tab:si-mkt-lin-500-beam-sweep}
\end{table}

\end{document}